\documentclass{ifacconf}

\usepackage{graphicx}      
\usepackage{natbib}        

\usepackage{amssymb}            
\usepackage{mathtools}          
\usepackage{mathrsfs}           

\mathtoolsset{showonlyrefs}     
\usepackage{subcaption}         
\usepackage{url}                
\usepackage{mathtools} 
\usepackage{wrapfig}
\usepackage{paralist}
\usepackage{siunitx}
\usepackage{placeins}

\DeclarePairedDelimiter{\norm}{\lVert}{\rVert} 
\newcommand{\fakepar}[1]{\vspace{1mm}\noindent\textbf{#1.}}
\newcommand{\ie}{i\/.\/e\/.,\/~}

\newcommand{\cf}{cf\/.\/~}

\newcommand{\abs}[1]{|#1|}

\begin{document}
\begin{frontmatter}

\title{CHEQ-ing the Box: Safe Variable Impedance Learning for Robotic Polishing\thanksref{footnoteinfo}} 

\thanks[footnoteinfo]{This work was funded by the German Federal Ministry of Education and Research (ProKI-Netz, grant number 02P22A010). Computations were performed with resources by RWTH Aachen University under the projects $thes1735$, $p0022348$, $p0022301$, and $p0021919$.}

\author[First]{Emma Cramer} 
\author[First]{Lukas Jäschke} 
\author[First]{Sebastian Trimpe}

\address[First]{Institute for Data Science in Mechanical Engineering\\
    RWTH Aachen University, Germany}

\begin{abstract}
Robotic systems are increasingly employed for industrial automation, with contact-rich tasks like polishing requiring dexterity and compliant behavior. These tasks are difficult to model, making classical control challenging. Deep reinforcement learning (RL) offers a promising solution by enabling the learning of models and control policies directly from data. However, its application to real-world problems is limited by data inefficiency and unsafe exploration.
Adaptive hybrid RL methods blend classical control and RL adaptively, combining the strengths of both: structure from control and learning from RL. This has led to improvements in data efficiency and exploration safety. However, their potential for hardware applications remains underexplored, with no evaluations on physical systems to date. Such evaluations are critical to fully assess the practicality and effectiveness of these methods in real-world settings. This work presents an experimental demonstration of the hybrid RL algorithm CHEQ for robotic polishing with variable impedance, a task requiring precise force and velocity tracking.
In simulation, we show that variable impedance enhances polishing performance. We compare standalone RL with adaptive hybrid RL, demonstrating that CHEQ achieves effective learning while adhering to safety constraints. On hardware, CHEQ achieves effective polishing behavior, requiring only eight hours of training and incurring just five failures. These results highlight the potential of adaptive hybrid RL for real-world, contact-rich tasks trained directly on hardware.
\end{abstract}

\begin{keyword}
reinforcement learning, variable impedance, robotic polishing
\end{keyword}

\end{frontmatter}
%
%
%
%
%
\section{Introduction}
Robotics has become a cornerstone of industrial automation, playing a pivotal role in improving efficiency and precision across various domains. A long-term aspiration in robotics is to achieve manipulation capabilities that combine human-like dexterity with compliant behavior, enabling robots for contact-rich tasks. In recent years, learning-based approaches, such as deep reinforcement learning (RL), have emerged to learn complex nonlinear control policies directly from data. While RL has shown great success in challenging control problems such as gameplay ~\citep{mnih_human-level_2015, silver_general_2018} and robotic manipulation ~\citep{buchler_learning_2022}, its application to real-world problems remains limited mainly due to its data inefficiency and unstructured exploration behavior. Prior work often focuses on sim-to-real transfer to mitigate these challenges. However, for highly complex or contact-rich tasks, it is often impractical to design accurate simulations that capture all necessary dynamics and interactions. As a result, direct learning on hardware sometimes becomes unavoidable. This introduces a critical challenge: the reliance of RL on random exploration can lead to unsafe behaviors, risking damage to both the robot and its environment. Addressing safe exploration is essential to enable safe and efficient RL training on hardware.

A prime reason for the random exploration behavior is the task-agnostic architecture of state-of-the-art RL approaches~\citep{haarnoja_soft_2018}, which lack the incorporation of prior knowledge on how to solve the task at hand. 
In contrast, control theory offers a rich set of methods for deriving near-optimal controllers based on first-principles modeling. These methods provide a decent baseline that works well when the system dynamics can be accurately modeled but fall short when the dynamics are complex or poorly understood.
This motivates hybrid RL~\citep{silver_residual_2018, johannink_residual_2019}, blending control priors with deep RL policies. Hybrid algorithms thus combine the informed behavior of the prior with task-specific RL optimization to tackle complex, nonlinear problems.

The majority of prior work in hybrid RL~\citep{silver_residual_2018, johannink_residual_2019, schoettler_deep_2020, ceola_resprect_2024} proposes a fixed weighting between control prior and RL agent. A fixed blending, however, disregards that the capability of the RL agent depends on training time and state. As more data is collected, the agent refines its behavior, ultimately outperforming the prior across larger portions of the task domain. Adaptive hybrid RL (AHRL) methods~\citep{cheng_control_2019, rana_bayesian_2023} adapt the weighting between RL agent and control prior based on the agent's confidence. The newly developed CHEQ algorithm~\citep{cramer_contextualized_2024} dynamically adjusts the weighting based on the parametric uncertainty of a critic ensemble. This has demonstrated significantly safer exploration behavior, \ie fewer violations of safety limits, and accelerated learning than traditional model-free RL, residual RL, and prior work in AHRL. 
The properties of AHRL methods, in general, and CHEQ, in particular, are promising for real-world applications and direct training on hardware. Hardware evaluations are critical, as real-world challenges like contact dynamics, friction, and state-dependent noise are difficult to simulate accurately. While some fixed-weight hybrid RL methods have trained on hardware~\citep{johannink_residual_2019, schoettler_deep_2020}, the advancements of AHRL have been validated exclusively in simulation and with low-dimensional action spaces. We argue that the full potential and the true challenges of these algorithms can only be revealed through evaluation on a challenging hardware problem.

In this work, we focus on polishing a 3D object with a robotic arm, a task that exemplifies the challenges of contact-rich manipulation in industrial automation. Polishing requires continuous contact with a surface while adhering to specific force and velocity profiles, demanding precise control to ensure quality and consistency. Variable impedance control (VIC) is particularly well-suited for such tasks~\citep{martin-martin_variable_2019, bogdanovic_learning_2020}, as it enables robots to dynamically adapt their stiffness and damping to interact effectively with their environment. However, defining suitable variable impedance gains is inherently complex, and manual tuning becomes impractical in scenarios with high-dimensional or nonlinear dynamics. This makes robotic polishing a compelling benchmark for evaluating RL-based control strategies.

This work is the first to apply AHRL, specifically the CHEQ algorithm, to a hardware learning problem. The hybrid RL agent outputs the impedance gains and the end-effector position and orientation. This output is then fed into a cascaded impedance controller. The goal is to (i) investigate the potential of CHEQ for learning variable impedance gains and (ii) explore its potential to learn such a policy safely, directly on hardware.

In simulation, we demonstrate that VIC significantly improves polishing behavior. We compare standard RL with AHRL, showing that CHEQ can learn effective impedance gains while maintaining exploration within safety limits. While CHEQ cannot provide theoretical safety guarantees, the control prior guides the exploration behavior to be close to the desired task. Finally, we deploy CHEQ on hardware and show that good polishing behavior can be achieved within eight hours and only five failures, underscoring its efficiency and safety in real-world scenarios.
%
%
%
%
%
\section{Related Work}
\label{sec:related_work}
In this section, we discuss relevant prior work in hybrid RL and works focusing on RL-based VIC and polishing.

\fakepar{Hybrid Reinforcement Learning}
Hybrid RL combines RL and control priors, categorized into fixed or adaptive weighting.

\citet{silver_residual_2018, johannink_residual_2019} first combined RL and control, introducing the term residual RL. In this work, we use the more general term hybrid RL to include approaches that adapt the controller's weight.~\citet{silver_residual_2018} and~\citet{johannink_residual_2019} show advantages of hybrid RL, such as sample efficiency and improved sim-to-real transfer. Fixed weight hybrid RL has been applied to tasks such as real robot insertion tasks~\citep{schoettler_deep_2020, davchev_residual_2022} and residual grasping policies on top of a simulation trained policy~\citep{ceola_resprect_2024}. A fixed mixing, however, does not consider the improving capabilities of the RL agent.

In contrast, AHRL adjusts weighting based on RL agent capabilities. ~\citet{rana_bayesian_2023} use a policy ensemble to estimate how certain the RL agent is in the current action. The combined action reflects the Bayesian posterior of control prior and policy distribution.~\citet{cheng_control_2019} use the TD-error as an uncertainty estimate and combine controller and RL agent based on this. ~\citet{rana_bayesian_2023} and~\citet{cheng_control_2019} train based on the combined action, which becomes brittle when facing large distributional shifts. ~\citet{cramer_contextualized_2024} demonstrate that the weighting factor induces a contextualized MDP, which must be observable by the agent. They train based on the RL action, incorporating the weighting factor into the agent's state. This approach achieves significantly safer exploration and fewer failures than standard RL, fixed weighting, and prior adaptive methods. This makes it an ideal AHRL method to investigate safe exploration during training on real hardware, which we do herein.

\fakepar{Reinforcement Learning-based VIC and Polishing}
Reinforcement Learning can be used to learn VIC, first demonstrated by~\citet{buchli_learning_2011}. Some works have focused on learning VIC from data gathered on hardware.~\citet{luo_reinforcement_2019} combine a higher model-based RL controller with a lower-level impedance controller, showing that this approach can learn real-world insertion tasks data efficiently. In~\citep{roveda_model-based_2020}, a model-based RL approach is trained to model human-robot interaction dynamics, and then an MPC is used to optimize VIC parameters.~\citet{anand_data-efficient_2024} learn VIC with a model-based RL agent through real-world interactions. They evaluate their approach on water-pouring with a robotic arm. While prior work emphasizes data-efficient model-based RL, both model-based and model-free algorithms are commonly used in RL. Here, we use model-free hybrid RL to remain safe during exploration.

Some prior work specifically focused on learning VIC for robotic polishing.~\citet{martin-martin_variable_2019} investigated vision-based surface wiping. While we aim to track a specific target force, they want to achieve an arbitrary force sufficient to wipe dirt. They trained in simulation and then transferred the policy to a real robot. ~\citet{bogdanovic_learning_2020} examined the simulated task of a circular end-effector motion while in contact with a table and applying a desired constant vertical force.~\citet{anand_evaluation_2022} consider wiping a 2D surface with a robotic arm, demanding force tracking in the vertical direction and motion tracking in the remaining five DOFs. The model-based RL agent tunes the two-dimensional impedance gains, while motion tracking is handled by a cascaded controller. This approach was directly trained on hardware, focusing on data efficiency. All approaches above consider wiping a 2D surface, which substantially reduces the action space of the RL agent. In this work, we want to use a more complex 3D task that requires adapting impedance gains for all Cartesian axes and orientations.
Both~\citet{martin-martin_variable_2019} and \citet{bogdanovic_learning_2020} investigate different choices of action spaces and conclude that a cascaded RL approach with variable impedance in end-effector space is most suited for the considered wiping task. We thus use this approach as our control pipeline and investigate the potential of AHRL for training directly on hardware.
%
%
%
%
%
\section{Robotic Polishing}
\label{sec:polishing}
\begin{figure}[tb]
     \centering
     \hspace{0.4cm}
     \begin{subfigure}[c]{0.29\columnwidth} 
         \centering
        \includegraphics[width=\textwidth]{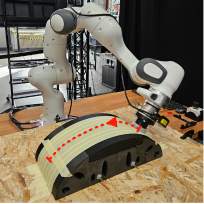}
            \vspace{-0.35cm}
     \end{subfigure}
     \hspace{0.4cm}
     \begin{subfigure}[c]{0.42\columnwidth} 
         \centering
        \includegraphics[width=\textwidth]{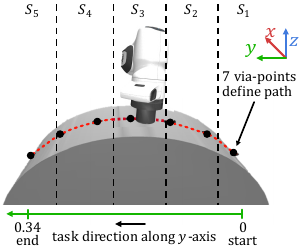}
        \vspace{-0.35cm}
     \end{subfigure}
     \hspace{0.4cm}
    \vspace{-0.25cm}
    \caption[format=plain]{Real robot setup and task description.}
    \label{fig:env_traj}
\end{figure}
We explore robotic polishing as a continuous contact force control task solvable with VIC. A non-rotating polishing tool is attached to a 7-DOF Franka Panda robotic arm, with a 3D-printed bridge as workpiece. We choose this object for its multiple challenges. The 3D movement showcases the need for variable impedance. Further, the task's movement direction does not align with Cartesian axes, and the workpiece features curvature variation along the y-axis, adding complexity. 

The task involves polishing along a path $r^\mathrm{path}$ defined via $7$ distinct via-points. Figure \ref{fig:env_traj} shows the simulated and real robot setup and the desired polishing path.
This research aims to achieve fine material removal through precise, low-force polishing, maintaining consistent material removal. Assuming a constant contact area, the material removal rate~\citep{fiedler_processing_1998} is $\frac{dh}{dt} = c \cdot F_{N} \cdot v_T$, 
with material thickness $h$, normal force on the workpiece $F_{N}$, translational velocity $v_T$ along $r^\mathrm{path}$, and $c$ a proportionality constant. Consistency in material removal rate thus requires maintaining constant force and translational velocity throughout the task.
%
%
%
%
%
\section{Learning Variable Impedance via CHEQ}
\begin{figure*}[tb]
     \centering
     \begin{subfigure}[c]{78mm}
         \centering
        \includegraphics[clip, trim=0cm 0.0cm 0cm 0cm]{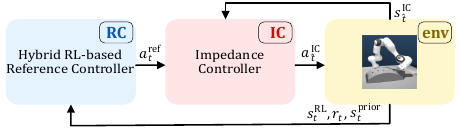}
        \vspace{-0.25cm}
         \caption{}
         \label{fig:polish_pipeline_overview}
     \end{subfigure}
     \begin{subfigure}[c]{70mm}
         \centering
        \includegraphics[clip, trim=0cm 0.0cm 0cm 0cm]{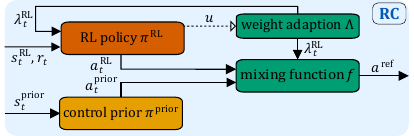}
        \vspace{-0.25cm}
         \caption{}
         \label{fig:cheq_overview}
     \end{subfigure}
    \vspace{-0.25cm}
     \caption{Panel (\subref{fig:polish_pipeline_overview}) shows the complete control pipeline with higher-level reference controller RC and lower-level impedance controller IC. Panel (\subref{fig:cheq_overview}) gives a detailed view of the RC, an adaptive hybrid RL agent based on CHEQ.}
     \label{fig:cheq}
\end{figure*}

Our control pipeline (\cf Fig. \ref{fig:cheq}) comprises a hybrid RL-based reference controller (RC) and cascaded impedance controller in end-effector space (IC)~\citep{khatib_inertial_1995}. 
The RC is based on our AHRL algorithm, CHEQ. This controller outputs $a_t^\mathrm{ref}$, including the next Cartesian end-effector pose and variable impedance gains. 
The IC outputs $a^\mathrm{IC}_{\hat{t}}$, mapping end-effector pose error to desired joint torques while keeping compliant end-effector behavior. 

\subsection{Impedance Controller (IC)}
We specify the robot end-effector pose $p \in \mathbb{R}^6$ by its position $x= (x_x, x_y, x_z)$ in Cartesian space and orientation $\phi=(\phi_x, \phi_y, \phi_z)$ in Euler angles. The complete end-effector pose can be written as $p =(x, \phi)$.  
We denote the pose error between the target and the current pose at time $t$ as $\Delta p = p^\mathrm{ref} - p$.
End-effector velocity and acceleration are defined as $\Dot{p}$ and $\Ddot{p}$ and their errors as $\Delta \Dot{p}$ and $\Delta \Ddot{p}$ respectively. We denote the joint space configuration by $q \in \mathbb{R}^7$.
The goal of impedance control is to model the behavior of a mass-spring-damper system in task space, such that $M \Delta \Ddot{p} + D \Delta \Dot{p} + K \Delta p = f_\mathrm{ext}$.
Here $M$, $D$, and $K$ $\in \mathbb{R}^{6\times6}$ represent the Cartesian inertia, damping, and stiffness matrices, and $f_\mathrm{ext}$ is a force applied to the robot. Commonly, $M=0$ is assumed to mitigate noisy acceleration measurements~\citep{lynch_modern_2017}. Moreover, we set $\Dot{p}^\mathrm{ref} = 0$. Under these assumptions, the desired joint torques to achieve this impedance behavior can be calculated as $\tau_\mathrm{IC}(q) = J^\intercal (K \Delta p - D \Delta \Dot{p})$, where $J$ is the Jacobian matrix. 
The controller also applies a null-space torque $\tau_\mathrm{null}(q)$, Coriolis and centrifugal forces $C(q,\Dot{q})$, and gravity compensation $g(q)$. The IC then outputs $a^\mathrm{IC}_{\hat{t}} = \tau_\mathrm{IC}(q_{\hat{t}}) + \tau_\mathrm{null}(q_{\hat{t}}) + C(q_{\hat{t}},\Dot{q}_{\hat{t}}) + g(q_{\hat{t}})$.

\subsection{Hybrid RL-based Reference Controller (RC)}
\label{sec:RC}
We use the CHEQ algorithm~\citep{cramer_contextualized_2024} as the RC, adaptively combining actions from the RL agent and nominal controller.

\fakepar{RL Agent}
We model the environment as a discounted Markov decision process defined by the tuple $\mathcal{M} = (\mathcal{S}, \mathcal{A}, p, r, \rho_0, \gamma)$, with state space $\mathcal{S}$, action space $\mathcal{A}$, start state distribution $\rho_0$ and transition function $p(s_{t+1}^\mathrm{RL}, r_{t+1} \mid s_t^\mathrm{RL}, a_t^{\mathrm{RL}})$. During transitions, rewards $r_t \in \mathbb{R}$ are emitted. The objective is to learn a policy $\pi^{\mathrm{RL}}(a^{\mathrm{RL}}_t \mid s_t^\mathrm{RL})$ that maximizes the expected cumulative sum of rewards discounted by $\gamma \in (0, 1)$. This objective can be written as
$J(\pi^\mathrm{RL}) = \max_{\pi^\mathrm{RL}} \mathbb{E}_{\pi^\mathrm{RL}, \mathcal{M}} \left[ \sum_{t=0}^{\infty} \gamma^t r_{t+1} \right]$. The action value or Q-function $Q^{\pi^\mathrm{RL}}(s_t^\mathrm{RL}, a^{\mathrm{RL}}_t)$ conditions expected return on specific state action pairs.

CHEQ builds upon the popular model-free soft-actor critic (SAC) architecture~\citep{haarnoja_soft_2018} and combines this with an ensemble of $E$ critic networks.
To reduce the action space, we define the stiffness matrix as a diagonal matrix $K=\mathrm{diag}(k_x, k_y, k_z, k_{\phi_x}, k_{\phi_y}, k_{\phi_z})$ and the damping by a damping factor $\zeta$ proportional to $K$ such that $D = 2 \zeta \cdot \sqrt{K}$. 
The agent then outputs 
\begin{equation}
    a^\mathrm{RL}_t = (\Delta p^\mathrm{RL}_t, K_t, \zeta_t).
\end{equation}
For the state space of the agent, we define $\Tilde{p}=(x, \phi^\mathrm{quat}$) where $\phi^\mathrm{quat} \in \mathbb{R}^4$ describes the orientation in unit quaternions. The complete state space can then be written as
\begin{equation}
    s^\mathrm{RL}_t=(q_t, \cos(q_t), \sin(q_t), \dot{q_t}, \Tilde{p_t}, \dot{\Tilde{p_t}}, F_t, \Delta r^\mathrm{path}_t) \in \mathbb{R}^{74}.
\end{equation}
Here, $F_t \in \mathbb{R}^3$ is the contact force of the end-effector, and $\Delta r^\mathrm{path}_t \in \mathbb{R}^{5\times6}$ describes the position and velocity errors of the end-effector to the next five points along the path $r^\mathrm{path}$. The contact force is sensed via a force-torque sensor.

At each time step $t$, the reward is a weighted sum of terms designed to encourage path-following, as well as velocity and force tracking such that
\begin{equation}
        r_t = c_{c^{\bot}} r_{c^{\bot},t} + c_{c^{\parallel}} r_{c^{\parallel},t} + c_d r_{d,t} + c_v r_{v,t} + c_f r_{f,t}.
\end{equation}
Here, $r_{c^{\bot}}$ and $r_{c^{\parallel}}$ reward path-following in perpendicular and parallel path direction, $r_d$ directional alignment, and $r_v$ and $r_f$ velocities and forces close to the respective targets. Each term is weighted with a constant $c$. In addition, we give a penalty $r_{trunc}$ for episode truncation and a final termination reward $r_{term}$ when all via-points are successfully wiped. Details are in~\citet{cramer_cheqingbox_2024}.

\fakepar{Nominal Control}
The nominal controller outputs $a^\mathrm{prior}_t$, including next desired position and orientation and fixed predefined impedance gains
\begin{equation}
    a^\mathrm{prior}_t= (\Delta p^\mathrm{prior}_{t}, K^\mathrm{prior}, \zeta^\mathrm{prior}).
\end{equation}
Following a simple and established approach, we define our reference path $r^\mathrm{path}$ by interpolating between 7 predefined via-points for the position~\citep{lynch_modern_2017} and using the SLERP method~\citep{shoemake_animating_1985} for the orientation. Since our hybrid controller might be far from the defined path, we cannot define the motion control with a fixed velocity. Instead, we choose the next control point $p^\mathrm{prior, ref}_{t}$ as the farthest point within a predefined radius.

\fakepar{CHEQ} CHEQ adaptively blends the action of the RL agent $a^\mathrm{RL}_t$ and the nominal controller $a^\mathrm{prior}_t$ (\cf Fig. \ref{fig:cheq_overview}).
The combined action $a^\mathrm{ref}_t$ is then a linear weighted sum
\begin{equation}
    a^\mathrm{ref}_t = (1-\lambda^\mathrm{RL}_t) \cdot a^\mathrm{prior}_t + \lambda^\mathrm{RL}_t \cdot a^\mathrm{RL}_t,
    \label{eq:apr-mixing-function}
\end{equation}
with the adaptive weight $\lambda^\mathrm{RL}_t \in [0, 1]$. 
The weight is adapted based on the uncertainty $u(s^\mathrm{RL}_t,a^\mathrm{RL}_t, \lambda^\mathrm{RL}_t)$ of an ensemble of $E$ Q-networks $Q_{\theta_e} \text{, with }e=1\hdots E$. To simplify notation we introduce $z_t = (s^\mathrm{RL}_t,a^\mathrm{RL}_t, \lambda^\mathrm{RL}_t)$. The uncertainty is estimated as the standard deviation of the $E$ critic predictions $u(z_t) = (\frac{1}{E} \sum_{e=1}^{E} \left( Q_{\theta_e}(z_t) - \mu (z_t) \right)^2)^{-1}$ with $\mu(z_t) =  \frac{1}{E} \sum_{e=1}^{E} Q_{\theta_e}(z_t)$.
The weighting factor $\lambda^\mathrm{RL}_t$ is then a clipped linear function between a minimum uncertainty $u_{\mathrm{min}}$ and maximum uncertainty $u_{\mathrm{max}}$.
%
%
%
%
%
\section{Simulation Study}
We first conducted a simulation study to show that (i) variable gains benefit the polishing task and that (ii) CHEQ can learn such gains with safe and efficient exploration.

\subsection{Evaluating the Need for VIC}
\begin{figure*}[tb]
     \centering
        \begin{subfigure}[b]{\textwidth}
        \begin{flushright}
        \includegraphics[clip, trim=0.1cm 0.1cm 0.1cm 0.1cm]{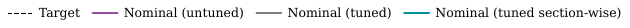}
        \end{flushright}
     \end{subfigure} 
     \begin{subfigure}[b]{0.32\textwidth}
         \centering
        \includegraphics[clip, trim=0cm 0.25cm 0cm 0.25cm]{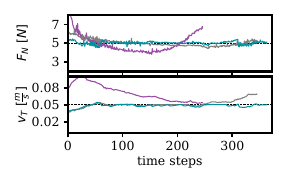}
        \vspace{-0.15cm}
         \caption{Force-velocity behavior.}
         \label{fig:fv_behavior_raw_tuned_partial}
     \end{subfigure}
     \begin{subfigure}[b]{0.32\textwidth}
         \centering
        \includegraphics[clip, trim=0cm 0.25cm 0cm 0.25cm]{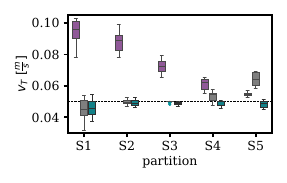}
        \vspace{-0.15cm}
         \caption{Velocity behavior per section.}
         \label{fig:v_box_raw_tuned_partial}
     \end{subfigure}
     \begin{subfigure}[b]{0.32\textwidth}
         \centering
        \includegraphics[clip, trim=0cm 0.25cm 0cm 0.25cm]{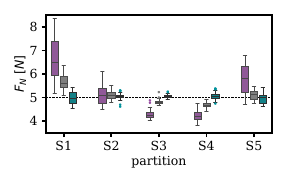}
        \vspace{-0.15cm}
         \caption{Force behavior per section.}
         \label{fig:f_box_raw_tuned_partial}
     \end{subfigure}
    \vspace{-0.25cm}
    \caption{Comparison of force and velocity over one episode~(\subref{fig:fv_behavior_raw_tuned_partial}) and boxplots for the five sections (\subref{fig:v_box_raw_tuned_partial},\subref{fig:f_box_raw_tuned_partial}). The fixed-gain controller tuned with BO shows increased performance, and tuning per section further improves behavior.}
    \label{fig:vic_raw_tuned_partial}
\end{figure*}
To show the benefits of variable impedance, we evaluate the nominal controller with fixed and partly fixed gains.

\fakepar{Experimental Setup} 
We define a fixed gain controller with suitable but untuned gains as a baseline, calling this controller \textit{Nominal (untuned)}. Using Bayesian Optimization (BO) in Weights\&Biases~\citep{wandb}, we tuned this controller over \num{500} episodes and refer to it as \textit{Nominal (tuned)}. 
Next, we partitioned the workpiece into \num{5} sections (see Fig. \ref{fig:env_traj}) and applied BO to optimize individual gains per section. This involved \num{500} episodes per section, iteratively using the best-found gains from previous sections. We call this \textit{Nominal (tuned section-wise)}.

\fakepar{Results}
Fig. \ref{fig:vic_raw_tuned_partial} compares the force-velocity behavior of the three controllers over an episode, along with boxplots for each of the five object sections. The dotted line marks the force and velocity targets of \qty{5}{\newton} and \qty{0.05}{\metre \per \second}. The untuned controller fails to meet the targets, while the BO-tuned fixed gain controller performs significantly better but shows larger errors at the task's beginning and end, where the object curves steeply. The section-wise tuned controller effectively reduces errors in these regions, showing that variable impedance gains can yield superior performance. However, dividing the controller into arbitrary sections is impractical with manual methods. RL offers the potential to tune state-dependent gains.

\subsection{Learning VIC with RL}
Next, we evaluate the effectiveness of RL in learning state-dependent VIC. Our findings show that CHEQ not only enables safe exploration during training but also successfully learns effective polishing. In contrast, standard RL only learns when unsafe behavior is permitted.
\begin{figure*}[tb]
    \centering
    \begin{subfigure}[b]{\textwidth}
        \begin{flushright}
        \includegraphics[clip, trim=0.1cm 0.1cm 0.1cm 0.1cm]{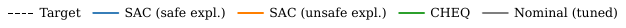}
        \end{flushright}
    \end{subfigure}
    \begin{minipage}[t]{\textwidth}
    \begin{subfigure}[b]{0.32\textwidth} 
        \centering
        \includegraphics[clip, trim=0cm 0.25cm 0cm 0.25cm]{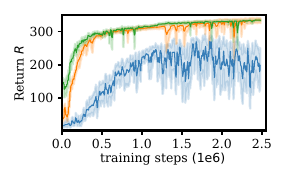}
        \vspace{-0.2cm}
        \caption{Return.}
        \label{fig:return_cheq_sac}
    \end{subfigure}
    \begin{subfigure}[b]{0.32\textwidth}
        \centering
        \includegraphics[clip, trim=0cm 0.25cm 0cm 0.25cm]{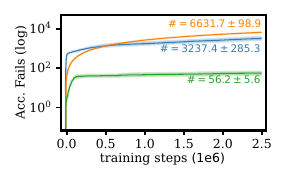}
        \vspace{-0.2cm}
        \caption{Accumulated failures (log scale).}
        \label{fig:fails_cheq_sac}
    \end{subfigure}
    \begin{subfigure}[b]{0.32\textwidth}
         \centering
        \includegraphics[clip, trim=0cm 0.25cm 0cm 0.25cm]{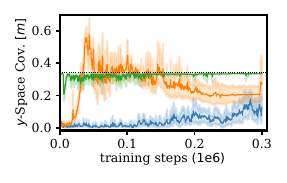}
        \vspace{-0.2cm}
         \caption{$y$-space coverage.}
         \label{fig:space_cov_cheq_sac}
     \end{subfigure}
     \end{minipage}
    \begin{minipage}[t]{\textwidth}
    \vspace{0.03cm}
    \begin{subfigure}[b]{0.32\textwidth}
        \centering
        \includegraphics[clip, trim=0cm 0.25cm 0cm 0.25cm]{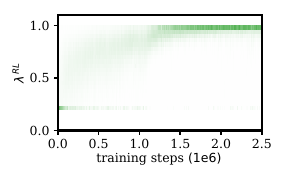}
        \vspace{-0.2cm}
        \caption{Weighting factor $\lambda^\mathrm{RL}$ distribution.}
        \label{fig:cheq_sim_lambda_dist}
    \end{subfigure}
    \begin{subfigure}[b]{0.32\textwidth}
        \centering
        \includegraphics[clip, trim=0cm 0.25cm 0cm 0.25cm]{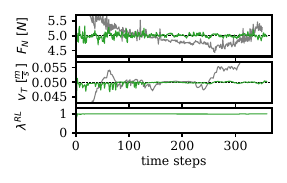}
        \vspace{-0.2cm}
        \caption{Force-velocity-lambda behavior.}
        \label{fig:fvw_behavior_cheq_tuned}
    \end{subfigure}
    \begin{subfigure}[b]{0.32\textwidth}
        \centering
        \includegraphics[clip, trim=0cm 0.25cm 0cm 0.25cm]{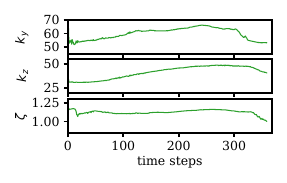}
        \vspace{-0.2cm}
        \caption{Impedance gains.}
        \label{fig:cheq_sim_impedance}
    \end{subfigure}
    \end{minipage}
    \vspace{-0.3cm}
    \caption{Comparison of return, fails and $y$-space coverage of CHEQ and SAC variants(\subref{fig:return_cheq_sac}-\subref{fig:space_cov_cheq_sac}). CHEQ achieves highest performance with only \num{56} fails. Over training, the distribution of the weighting factor $\lambda^\mathrm{RL}$ rises to \num{1} (\subref{fig:cheq_sim_lambda_dist}). 
    Fig. \subref{fig:fvw_behavior_cheq_tuned} and \subref{fig:cheq_sim_impedance} show the trained agent after \num{2.5e6} steps. The agent achieves good polishing behavior by adapting the gains.} 
    \label{fig:results_sim_cheq_sac}
\end{figure*}

\fakepar{Experimental Setup} 
We compared two SAC variants, \textit{SAC (unsafe exploration)} and \textit{SAC (safe exploration)}, along with our hybrid \textit{CHEQ} method. The latter two enforce safe exploration by applying velocity and force limits, bounding the end-effector's position, and constraining orientation to task-appropriate values (Details in \citet{cramer_cheqingbox_2024}). Episodes are truncated upon constraint violations to ensure minimal hardware wear and safeguard nearby humans. For \textit{SAC (unsafe exploration)}, all restrictions are lifted. The CHEQ agent uses the \textit{Nominal (untuned)} controller as its prior, arguing that an easy-to-define, untuned controller suffices for hybrid RL.

\fakepar{Results}
Figures \ref{fig:results_sim_cheq_sac}\subref{fig:return_cheq_sac}-\subref{fig:space_cov_cheq_sac} show the performance of RL agents during training, averaged over \num{10} runs with \qty{95}{\percent} quantiles. \textit{SAC (safe exploration)} achieves low returns, failing to learn the task effectively. In contrast, \textit{SAC (unsafe exploration)} learns the task but accumulates over \num{6e3} failures. We count only the first violation per episode, though actual failures are much higher, as we do not truncate episodes. This high failure rate makes it unsuitable for hardware due to excessive wear and tear.
The \textit{CHEQ} agent achieves high returns, slightly outperforming unsafe SAC while reducing failures by three orders of magnitude. Differences between approaches are evident in the $y$-space coverage (Fig. \ref{fig:space_cov_cheq_sac}). Since the polishing path is along the y-axis (see Fig. \ref{fig:env_traj}), $y$-space coverage is a proxy for task space exploration.
In the first \num{300e3} training steps, the safe SAC agent covers less than half the task, as safety constraints and random exploration limit progress. The unsafe SAC agent covers the full space but overshoots the endpoint. For CHEQ, the nominal controller guides exploration efficiently towards the full task space, and the small $\lambda^\mathrm{RL}$ at the beginning ensures the agent uses just enough random exploration to learn the task safely and efficiently.
Looking at the $\lambda^\mathrm{RL}$-distribution (see Fig. \ref{fig:cheq_sim_lambda_dist}), the agent first learns a small residual on top of the nominal controller. This already provides good polishing behavior and high returns. Over time, the agent transitions to full RL control, continuously improving performance. This induces a learning curriculum, ensuring safe exploration.
Compared to the tuned baseline (Fig. \ref{fig:fvw_behavior_cheq_tuned}), CHEQ achieves superior force and velocity tracking by adapting gains dynamically throughout the task (Fig. \ref{fig:cheq_sim_impedance}). A complete comparison with all priors is in \citet{cramer_cheqingbox_2024}. Overall, CHEQ demonstrates significantly safer learning than standard RL while outperforming the tuned baseline.
%
%
%
%
%
%
%
\section{Case Study: Real-world 3D polishing}
In this section, we address the challenges and solutions of training CHEQ on hardware, demonstrated through a 3D polishing task with a robotic manipulator. This task, representative of contact-rich manufacturing problems, marks CHEQ's first successful hardware implementation.
\begin{figure*}[tb]
    \centering
    \begin{subfigure}[b]{\textwidth}
        \begin{flushright}
        \includegraphics[clip, trim=0.1cm 0.1cm 0.1cm 0.1cm]{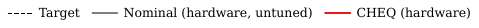}
        \end{flushright}
    \end{subfigure}
    \begin{subfigure}[b]{0.32\textwidth}
        \centering
        \includegraphics[clip, trim=0cm 0.25cm 0cm 0.25cm]{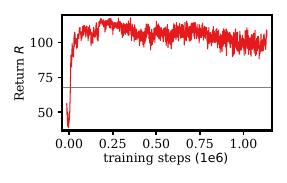}
        \vspace{-0.2cm}
        \caption{Return.}
        \label{fig:cheq_rr_rollout_return_long}
    \end{subfigure}
    \begin{subfigure}[b]{0.32\textwidth}
        \centering
        \includegraphics[clip, trim=0cm 0.25cm 0cm 0.25cm]{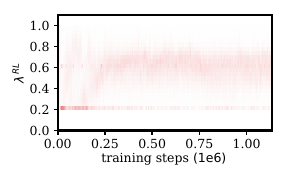}
        \vspace{-0.2cm}
        \caption{Weighting factor $\lambda^\mathrm{RL}$ distribution.}
        \label{fig:cheq_rr_lambda_dist}
    \end{subfigure}
    \begin{subfigure}[b]{0.32\textwidth}
        \centering
        \includegraphics[clip, trim=0cm 0.25cm 0cm 0.25cm]{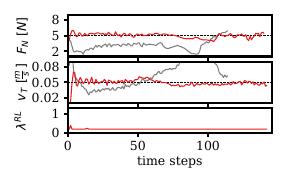}
        \vspace{-0.2cm}
        \caption{Force-velocity-lambda eval. ($\sim$\num{90e3})}
        \label{fig:fvw_behavior_cheq_rr_early}
    \end{subfigure}
    \vspace{-0.3cm}
    \caption{Return (\subref{fig:cheq_rr_rollout_return_long}) and $\lambda^\mathrm{RL}$-distribution (\subref{fig:cheq_rr_lambda_dist}) of CHEQ on hardware. The agent achieves good polishing behavior (\subref{fig:fvw_behavior_cheq_rr_early}).}
    \label{fig:rr_cheq}
\end{figure*}

\fakepar{Experimental Setup}
Additional challenges need to be considered to achieve stable training on hardware. On our hardware setup, we can achieve a maximum control frequency of \qty{20}{\hertz} while simultaneously dealing with measurement noise in the robot state and force measurements. Further, our simulation results revealed one critical aspect specific to CHEQ. In the initial training phase, the weighting factor fluctuates greatly between time steps, resulting in highly fluctuating IC gains. In simulation, this does not pose a problem. On real hardware, however, the gain fluctuations lead to a chattering end-effector motion. This doubly affects training as it complicates the control and increases the motion-dependent force sensor noise. To mitigate this, we use an average of \num{10} time steps to compute our uncertainty, leading to smoother weights. Combining the mentioned challenges, we find that data collected on our hardware is less reliable than in simulation. This might lead to unstable training. CHEQ provides a mechanism to mitigate this by slowing the curriculum. Thus, for our hardware experiments, we set more conservative uncertainty limits. This reduces the agency of the RL agent in the beginning and leads to more stable, safer, albeit slower, learning. We use an untuned prior (\textit{Nominal hardware (untuned)}) and set a UTD ratio of \num{2} to account for the higher cost of environment steps on hardware.

\fakepar{Results}
On hardware, the return improves quickly, surpassing the \textit{Nominal hardware (untuned)} controller after \num{6.5e3} steps. Even this early in the training process (\num{250e3} steps and \num{8} hours of training), good polishing behavior is achieved. The trained RC maintains smooth force and velocity close to the targets, achieving this with a low weight around \num{0.2} (see Fig. \ref{fig:fvw_behavior_cheq_rr_early}). Importantly, this behavior is achieved while accumulating only five failures.

Longer training does not further improve the return (\cf Fig. \ref{fig:cheq_rr_rollout_return_long}). In our adaptive setting, the agent must learn the complete state, action, and weight space. Whenever the agent learns a higher weight, the return drops and rises again once the agent has mastered this part. With the slow curriculum, the agent has not yet learned the complete weight space. Still, we see no benefit in further training, as our early results show that a small residual on the untuned controller suffices for effective polishing. Even though a residual term is enough to learn good polishing behavior in this setting, CHEQ still works differently than traditional fixed-weight approaches. \citet{cramer_contextualized_2024} demonstrated that a fixed weight compromises safe learning and performance; comparable final performance can only be achieved with the optimal weight for a certain scenario and control prior. However, even then, training benefits from the curriculum induced by adaptive weighting. We conclude that CHEQ learns good polishing behavior safely and efficiently on hardware.
%
%
%
%
%
%
\section{Conclusion}
This work applied the CHEQ algorithm, a novel AHRL algorithm, to robotic polishing\footnote{\scriptsize \url{github.com/Data-Science-in-Mechanical-Engineering/polishing-cheq}.}. In a simulation study, we highlighted the benefits of variable impedance and showed that CHEQ outperforms standalone RL. The adaption mechanism ensured safe exploration by gradually transitioning from control priors to autonomous operation. On hardware, CHEQ achieved effective polishing performance. This study demonstrates the potential of AHRL for hardware training, combining safety with performance. 

\begin{ack}
We thank Johannes Berger, Aditya Pradhan, Lukas Wildberger, and Furkan Celik for their help with the hardware implementations and input during the early stages of this research. We also thank Alexander Gräfe for the many helpful discussions. This work was funded in part by the German Federal Ministry of Education and Research (“Demonstrations- und Transfernetzwerk KI in der Produktion (ProKI-Netz)” initiative, grant number 02P22A010). Computations were performed with computing resources granted by RWTH Aachen University under the projects $thes1735$, $p0022348$, $p0022301$, and $p0021919$.
\end{ack}

\bibliography{references_pruned} 

\begin{thebibliography}{27}
\providecommand{\natexlab}[1]{#1}
\providecommand{\url}[1]{\texttt{#1}}
\providecommand{\urlprefix}{URL }
\expandafter\ifx\csname urlstyle\endcsname\relax
  \providecommand{\doi}[1]{doi:\discretionary{}{}{}#1}\else
  \providecommand{\doi}{doi:\discretionary{}{}{}\begingroup \urlstyle{rm}\Url}\fi

\bibitem[{Anand et~al.(2022)Anand, Hagen~Myrestrand, and Gravdahl}]{anand_evaluation_2022}
Anand, A.S., Hagen~Myrestrand, M., and Gravdahl, J.T. (2022).
\newblock Evaluation of {{Variable Impedance-}} and {{Hybrid Force}}/{{MotionControllers}} for {{Learning Force Tracking Skills}}.
\newblock In \emph{{{IEEE}} Int. Symp. on Syst. Integr.}, 83--89.

\bibitem[{Anand et~al.(2024)Anand, Kaushik, Gravdahl, and {Abu-Dakka}}]{anand_data-efficient_2024}
Anand, A.S., Kaushik, R., Gravdahl, J.T., and {Abu-Dakka}, F.J. (2024).
\newblock Data-{{Efficient Reinforcement Learning}} for {{Variable Impedance Control}}.
\newblock \emph{IEEE Access}, 12, 15631--15641.

\bibitem[{Biewald(2020)}]{wandb}
Biewald, L. (2020).
\newblock \urlprefix\url{https://www.wandb.com/}.

\bibitem[{Bogdanovic et~al.(2020)Bogdanovic, Khadiv, and Righetti}]{bogdanovic_learning_2020}
Bogdanovic, M., Khadiv, M., and Righetti, L. (2020).
\newblock Learning variable impedance control for contact sensitive tasks.
\newblock \emph{IEEE Rob. and Aut. Letters}, 5(4), 6129--6136.

\bibitem[{Buchli et~al.(2011)Buchli, Stulp, Theodorou, and Schaal}]{buchli_learning_2011}
Buchli, J., Stulp, F., Theodorou, E., and Schaal, S. (2011).
\newblock Learning variable impedance control.
\newblock \emph{Int. J. of Rob. Res.}, 30(7), 820--833.

\bibitem[{Büchler et~al.(2022)Büchler, Guist, Calandra, Berenz, Schölkopf, and Peters}]{buchler_learning_2022}
Büchler, D., Guist, S., Calandra, R., Berenz, V., Schölkopf, B., and Peters, J. (2022).
\newblock Learning to {Play} {Table} {Tennis} {From} {Scratch} {Using} {Muscular} {Robots}.
\newblock \emph{IEEE Trans. on Rob.}, 38(6), 3850--3860.

\bibitem[{Ceola et~al.(2024)Ceola, Rosasco, and Natale}]{ceola_resprect_2024}
Ceola, F., Rosasco, L., and Natale, L. (2024).
\newblock {RESPRECT}: {Speeding}-up {Multi}-{Fingered} {Grasping} {With} {Residual} {Reinforcement} {Learning}.
\newblock \emph{IEEE Rob. and Aut. Letters}, 9(4), 3045--3052.

\bibitem[{Cheng et~al.(2019)Cheng, Verma, Orosz, Chaudhuri, Yue, and Burdick}]{cheng_control_2019}
Cheng, R., Verma, A., Orosz, G., Chaudhuri, S., Yue, Y., and Burdick, J. (2019).
\newblock Control {Regularization} for {Reduced} {Variance} {Reinforcement} {Learning}.
\newblock In \emph{Int. Conf. on Mach. Learn.} PMLR.

\bibitem[{Cramer et~al.(2024{\natexlab{a}})Cramer, Frauenknecht, Sabirov, and Trimpe}]{cramer_contextualized_2024}
Cramer, E., Frauenknecht, B., Sabirov, R., and Trimpe, S. (2024{\natexlab{a}}).
\newblock Contextualized {Hybrid} {Ensemble} {Q}-learning: {Learning} {Fast} with {Control} {Priors}.
\newblock In \emph{Reinforcement {Learning} {Journal}}, volume~2, 926--945. Amherst Massachusetts.

\bibitem[{Cramer et~al.(2024{\natexlab{b}})Cramer, Jäschke, and Trimpe}]{cramer_cheqingbox_2024}
Cramer, E., Jäschke, L., and Trimpe, S. (2024{\natexlab{b}}).
\newblock {CHEQ}-ing the {Box}: {Safe} and {Efficient} {Variable} {Impedance} {Learning} for {Robotic} {Polishing} (extended version).
\newblock \urlprefix\url{http://tiny.cc/cramer_polishing_2024}.

\bibitem[{Davchev et~al.(2022)Davchev, Luck, Burke, Meier, Schaal, and Ramamoorthy}]{davchev_residual_2022}
Davchev, T., Luck, K.S., Burke, M., Meier, F., Schaal, S., and Ramamoorthy, S. (2022).
\newblock Residual {Learning} {From} {Demonstration}: {Adapting} {DMPs} for {Contact}-{Rich} {Manipulation}.
\newblock \emph{IEEE Rob. and Aut. Let.}, 7(2), 4488--4495.

\bibitem[{Fiedler(1998)}]{fiedler_processing_1998}
Fiedler, K.H. (1998).
\newblock Processing ({{Grinding}} and {{Polishing}}).
\newblock In \emph{The {{Properties}} of {{Optical Glass}}}, 245--262. Springer.

\bibitem[{Haarnoja et~al.(2018)Haarnoja, Zhou, Abbeel, and Levine}]{haarnoja_soft_2018}
Haarnoja, T., Zhou, A., Abbeel, P., and Levine, S. (2018).
\newblock Soft actor-critic: {Off}-policy maximum entropy deep reinforcement learning with a stochastic actor.
\newblock In \emph{Int. Conf. on Mach. Learn.} PMLR.

\bibitem[{Johannink et~al.(2019)Johannink, Bahl, Nair, Luo, Kumar, Loskyll, Ojea, Solowjow, and Levine}]{johannink_residual_2019}
Johannink, T., Bahl, S., Nair, A., Luo, J., Kumar, A., Loskyll, M., Ojea, J.A., Solowjow, E., and Levine, S. (2019).
\newblock Residual reinforcement learning for robot control.
\newblock In \emph{Int. Conf. on Rob. and Aut.({{ICRA}})}, 6023--6029. IEEE.

\bibitem[{Khatib(1995)}]{khatib_inertial_1995}
Khatib, O. (1995).
\newblock Inertial {{Properties}} in {{Robotic Manipulation}}: {{An Object-Level Framework}}.
\newblock \emph{Int. J. of Rob. Res.}, 14(1), 19--36.

\bibitem[{Luo et~al.(2024)Luo, Hu, Xu, Tan, Berg, Sharma, Schaal, Finn, Gupta, and Levine}]{luo_serl_2024}
Luo, J., Hu, Z., Xu, C., Tan, Y.L., Berg, J., Sharma, A., Schaal, S., Finn, C., Gupta, A., and Levine, S. (2024).
\newblock {{SERL}}: {{A Software Suite}} for {{Sample-Efficient Robotic Reinforcement Learning}}.

\bibitem[{Luo et~al.(2019)Luo, Solowjow, Wen, Ojea, Agogino, Tamar, and Abbeel}]{luo_reinforcement_2019}
Luo, J., Solowjow, E., Wen, C., Ojea, J.A., Agogino, A.M., Tamar, A., and Abbeel, P. (2019).
\newblock Reinforcement {{Learning}} on {{Variable Impedance Controller}} for {{High-Precision Robotic Assembly}}.
\newblock In \emph{Int. Conf. on Rob. and Aut.({{ICRA}})}, 3080--3087.

\bibitem[{Lynch and Park(2017)}]{lynch_modern_2017}
Lynch, K.M. and Park, F.C. (2017).
\newblock \emph{Modern Robotics: Mechanics, Planning, and Control}.
\newblock Cambridge University Press.

\bibitem[{{Mart{\'i}n-Mart{\'i}n} et~al.(2019){Mart{\'i}n-Mart{\'i}n}, Lee, Gardner, Savarese, Bohg, and Garg}]{martin-martin_variable_2019}
{Mart{\'i}n-Mart{\'i}n}, R., Lee, M.A., Gardner, R., Savarese, S., Bohg, J., and Garg, A. (2019).
\newblock Variable impedance control in end-effector space: {{An}} action space for reinforcement learning in contact-rich tasks.
\newblock In \emph{{{IEEE}} Int. Conf. on Int. Rob. and Sys. ({{IROS}})}.

\bibitem[{Mnih et~al.(2015)Mnih, Kavukcuoglu, Silver, Rusu, Veness, Bellemare, Graves, Riedmiller, Fidjeland, and Ostrovski}]{mnih_human-level_2015}
Mnih, V., Kavukcuoglu, K., Silver, D., Rusu, A.A., Veness, J., Bellemare, M.G., Graves, A., Riedmiller, M., Fidjeland, A.K., and Ostrovski, G. (2015).
\newblock Human-level control through deep reinforcement learning.
\newblock \emph{Nature}, 518(7540), 529.

\bibitem[{Rana et~al.(2023)Rana, Dasagi, Haviland, Talbot, Milford, and Sünderhauf}]{rana_bayesian_2023}
Rana, K., Dasagi, V., Haviland, J., Talbot, B., Milford, M., and Sünderhauf, N. (2023).
\newblock Bayesian controller fusion: {Leveraging} control priors in deep reinforcement learning for robotics.
\newblock \emph{Int. J. of Rob. Res.}, 42(3), 123--146.

\bibitem[{Roveda et~al.(2020)Roveda, Maskani, Franceschi, Abdi, Braghin, Molinari~Tosatti, and Pedrocchi}]{roveda_model-based_2020}
Roveda, L., Maskani, J., Franceschi, P., Abdi, A., Braghin, F., Molinari~Tosatti, L., and Pedrocchi, N. (2020).
\newblock Model-based reinforcement learning variable impedance control for human-robot collaboration.
\newblock \emph{J. of Int. Rob. Sys.}, 100(2), 417--433.

\bibitem[{Schoettler et~al.(2020)Schoettler, Nair, Luo, Bahl, Ojea, Solowjow, and Levine}]{schoettler_deep_2020}
Schoettler, G., Nair, A., Luo, J., Bahl, S., Ojea, J.A., Solowjow, E., and Levine, S. (2020).
\newblock Deep reinforcement learning for industrial insertion tasks with visual inputs and natural rewards.
\newblock In \emph{Int. Conf. on Int. Rob. and Sys({{IROS}})}, 5548--5555. IEEE.

\bibitem[{Shoemake(1985)}]{shoemake_animating_1985}
Shoemake, K. (1985).
\newblock Animating rotation with quaternion curves.
\newblock In \emph{Proceedings of the 12th Annual Conference on {{Computer}} Graphics and Interactive Techniques}, 245--254. Association for Computing Machinery, New York, NY, USA.

\bibitem[{Silver et~al.(2018{\natexlab{a}})Silver, Hubert, Schrittwieser, Antonoglou, Lai, Guez, Lanctot, Sifre, Kumaran, Graepel, Lillicrap, Simonyan, and Hassabis}]{silver_general_2018}
Silver, D., Hubert, T., Schrittwieser, J., Antonoglou, I., Lai, M., Guez, A., Lanctot, M., Sifre, L., Kumaran, D., Graepel, T., Lillicrap, T., Simonyan, K., and Hassabis, D. (2018{\natexlab{a}}).
\newblock A general reinforcement learning algorithm that masters chess, shogi, and {Go} through self-play.
\newblock \emph{Science}, 362(6419), 1140--1144.

\bibitem[{Silver et~al.(2018{\natexlab{b}})Silver, Allen, Tenenbaum, and Kaelbling}]{silver_residual_2018}
Silver, T., Allen, K., Tenenbaum, J., and Kaelbling, L. (2018{\natexlab{b}}).
\newblock Residual policy learning.
\newblock \emph{arXiv preprint arXiv:1812.06298}.

\bibitem[{Todorov et~al.(2012)Todorov, Erez, and Tassa}]{todorov_mujoco_2012}
Todorov, E., Erez, T., and Tassa, Y. (2012).
\newblock Mujoco: {A} physics engine for model-based control.
\newblock In \emph{{IEEE} Int. Conf. on Int. Rob. and Sys({{IROS}})}, 5026--5033. IEEE.

\end{thebibliography}
\appendix

%
%
%
%
%
\section{Experimental Details}    
This section gives further details on the polishing environment and the nominal controller, as well as hyperparameters for all experiments.

\subsection{Derivation of the Task Goal}
\label{sec:app_task_goal}
The goal of our task is to polish over the workpiece with a constant material removal rate (MRR). For this, we let a robotic arm with a polishing tool attached to the last joint follow a predefined path $r^\mathrm{path}$ over the bridge workpiece. To refine the requirements, further insights about the definition of MRR can be derived for our task setup.  

The normal pressure $p_N$ is defined as the normal force applied to a certain contact region of the surface $A \in \mathbb{R}$. In the case of low friction, the normed force is similar to the normal force $F_N$, as tangential forces can be omitted. When a spherical workpiece has contact with a plane polishing tool, the contact region has the geometry of an ellipsis. The size and shape of the ellipsis depend on the curvature of the workpiece, the applied force, and some material constants. Due to small expected variations in contact force, we approximate the area of contact to be constant.
For this assumption to be valid, we require alignment between tool orientation and workpiece surface normal.

When the tool is moving along the path $r^\mathrm{path}$, the relative velocity of the tool $v_{rel}$ can be split into two parts: translational velocity $v_T \in \mathbb{R}$ and rotational velocity $v_B \in \mathbb{R}$. As we use a non-rotating polishing tool, we assume $v_B = 0$. The translational velocity $v_T$ depicts the movement along the path. 

Given these insights, we can calculate the material removal rate as
\begin{equation}
  \begin{aligned}
    \frac{dh}{dt} &= K_p \cdot p_N \cdot v_{rel} \\
    &= K_p \cdot \frac{F_N}{A} \cdot (v_B + v_T) \\
    &= \frac{K_p}{A} \cdot F_N \cdot v_T \\
    \end{aligned}  
\end{equation}

The term $\frac{K_p}{A}$ is a constant factor. A constant MRR is therefore achieved if a constant force $F_N^\mathrm{target} \in \mathbb{R}$ and constant translational velocity $v_T^\mathrm{target} \in \mathbb{R}$ is maintained for the entire task. 
\subsection{Nominal Control Design}
\label{sec:app_nominal_control}

We use a classical motion control approach for the nominal controller~\citep{lynch_modern_2017}. This control approach can be separated into path generation and motion control.

As the action space is defined in task space, the path $r^\mathrm{path}$ is defined via Cartesian positions and orientations.
For the positions, we apply the Via-Points algorithm~\citep{lynch_modern_2017} using a third-degree interpolation method. The algorithm requires the specification of $N$ via-points with their corresponding positions $\tilde{x}_n \in \mathbb{R}^3$ and directional vectors $\tilde{d}_n \in \mathbb{R}^3$ of the path, for all $n \in N$ via-points. The directional vectors $ \tilde{d}_n$ are necessary hyperparameters for the Via-Points algorithm that represent the current path direction at the position of each via-point. Changing the directional vectors of a via-point leads to different curvatures of the resulting positional path. The position and the directional vectors then act as boundary constraints for the path, resulting in a well-defined analytical problem. The algorithm then outputs a piecewise function between each neighbouring via-point pair. 
For the orientation, we apply the spherical linear interpolation (SLERP) algorithm for quaternions~\citep{shoemake_animating_1985}. For SLERP, we specify the orientation of the end-effector $\tilde{\phi}_n \in \mathbb{R}^4$ at the positions of the $N$ via-points. SLERP then results in the definition of piecewise functions between each orientation specification $\tilde{\phi}_n$ and $\tilde{\phi}_{n+1}$.
After creation, the functions for the position and orientation are combined to form a coherent and continuous path function $r^\mathrm{path}$. This path is then discretized into $M \gg N$ equally spaced control points $p^\mathrm{prior, path}_{m} \in \mathbb{R}^7$.

\begin{figure}[ht]
    \centering
    \includegraphics[width=0.99\linewidth, trim=1cm 9.5cm 0cm 2.5cm]{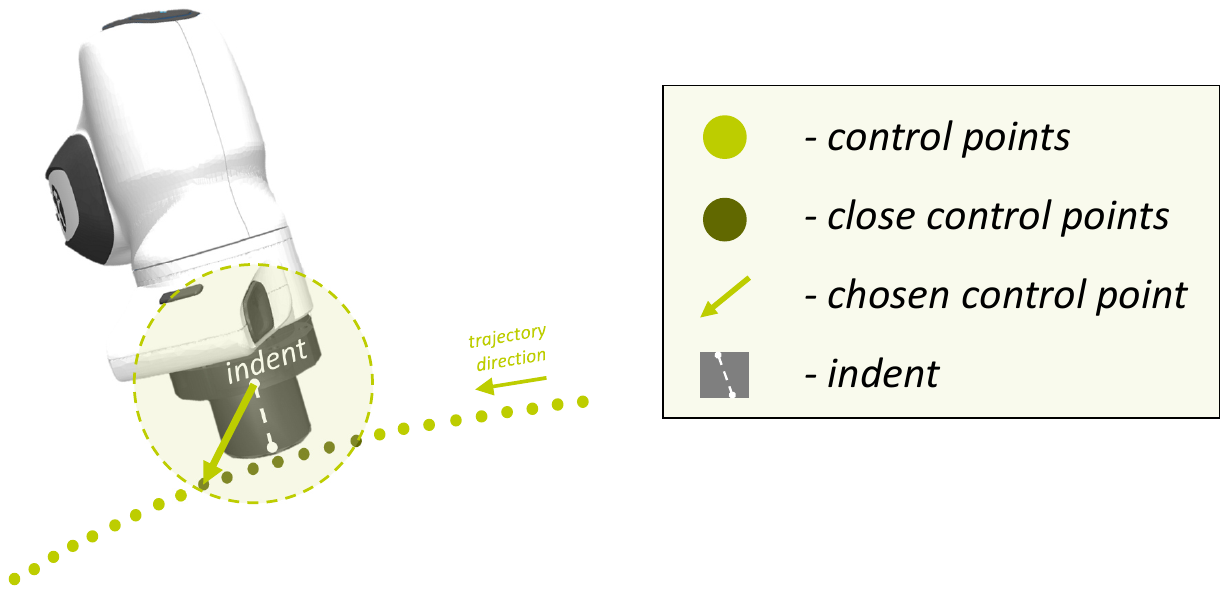}
    \caption{Generation of the next desired control point.}
    \label{fig:nominal_motion_law}
\end{figure}

The control points build the foundation for the motion control, which outputs the action $a^\mathrm{prior}_t$. 
Given the current pose $p_t = (x_t, \phi_t)^\intercal \in \mathbb{R}$ of the robotic arm, we search for close control points that lie within a predefined radius $a_r \in \mathbb{R}$. 
Close control points can be grouped in the set $P^\mathrm{close}$. From this set, we choose the control point $p^\mathrm{prior, ref}$ that is furthest along the task direction. The chosen control point is then used as a reference for the controller. Figure \ref{fig:nominal_motion_law} visualizes this process.

After determining the reference control point $p^\mathrm{prior, ref} = \left( x^\mathrm{prior, ref}, \phi^\mathrm{prior, ref} \right)$, the action $a^\mathrm{prior}_t$ is computed using the following two equations for position and orientation

\begin{align}
        \Delta x^\mathrm{prior}_t = x_t - x^\mathrm{prior, ref}, \\
        \Delta \phi^\mathrm{prior}_t = \phi^\mathrm{prior, ref} \cdot \phi_t^{-1}.
\end{align}

Since the positional part of the path is defined directly on the workpiece surface, we shift the robot's position $x_t$ to be located inside the end-effector. For this purpose, we define an indentation parameter $\Delta x^\mathrm{indent}$ that describes the absolute distance from position $x_t$ to the end-effector's surface. Changing the indentation modifies the force behavior, with higher values leading to a stronger movement toward the workpiece.

To receive the full nominal controller action $a^\mathrm{prior}_t$, $\Delta x^\mathrm{prior}_t$ and $\Delta \phi^\mathrm{prior}_t$ are concatenated into $\Delta p^\mathrm{prior}_{t}$ and enhanced with constant impedance gains $K^\mathrm{prior}, \zeta^\mathrm{prior}$. The full nominal controller action is then defined as
\begin{equation}
        a^\mathrm{prior}_t= (\Delta p^\mathrm{prior}_{t}, K^\mathrm{prior}, \zeta^\mathrm{prior}).
\end{equation}

\subsection{Details on the Reward Definition}
\label{sec:app_reward_def}
At each time step $t$, the reward is a weighted sum of terms designed to encourage path-following, as well as velocity and force tracking such that
\begin{equation}
        r_t = c_{c^{\bot}} r_{c^{\bot},t} + c_{c^{\parallel}} r_{c^{\parallel},t} + c_d r_{d,t} + c_v r_{v,t} + c_f r_{f,t}.
\end{equation}
Here, $r_{c^{\bot}}$ and $r_{c^{\parallel}}$ reward path-following in perpendicular and parallel path direction, $r_d$ directional alignment, and $r_v$ and $r_f$ velocities and forces close to the respective targets. Each term is weighted with a constant $c_i \in [0,1]$, and we choose these constants such that $\sum_{i} c_i = 1$. Since we normalize all reward terms, this choice of weighting factor makes it easy to interpret the relation between each reward term. In addition, we give a penalty $r_{trunc}$ for episode truncation and a final termination reward $r_{term}$ when the last via-point is successfully wiped. 

We use two piecewise functions to transform an error value $e \in \mathbb{R}^+$ of the current state $s_t$ into a reward. 
The function $l(e)$ is a purely linear mapping from error to reward and can be written as
\begin{equation*}
        l(e) = \begin{cases}
                    1 - \frac{e}{e_{max}}, & 0\leq e \leq e_{max} \\
                    0, & e > e_{max}.
                \end{cases}
\end{equation*}
The function $q(e)$ rewards values close to the targets quadratically and values far away from the targets linearly. The function is defined as
\begin{equation*}
        q(e) = \begin{cases}
                    1 - \frac{e}{2e_{mid}}, & 0\leq e \leq e_{mid} \\
                    \frac{e_{max}^2 - e^2}{2(e_{max}^2 - e_{mid}^2)}, & e_{mid} < e \leq e_{max} \\
                    0, & e > e_{max}. 
                \end{cases}
\end{equation*}
\begin{figure}[ht]
    \centering
    \includegraphics[width=0.99\linewidth, trim=3cm 9.5cm 0cm 2.5cm, clip]{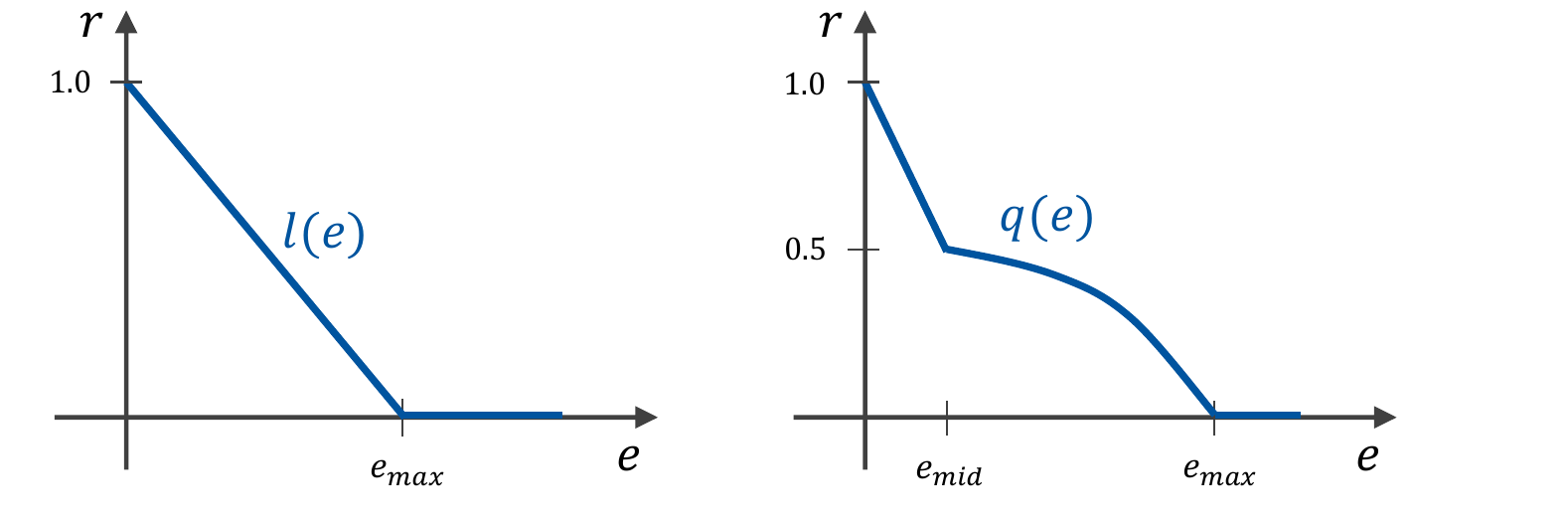}
    \caption{We define two functions to map an error to a reward. The left panel shows the linear transformation $l(e)$. The right panel shows the transformation $q(e)$ mapping outer values quadratically and values close to targets linearly.}
    \label{fig:piecewise}
\end{figure}
Figure \ref{fig:piecewise} visualizes the two mapping functions.

\fakepar{Path following}
The robot's path-following behavior is reinforced by using four reward terms 
\begin{compactenum}
    \item the cross-error $e_{c^{\parallel}}$ parallel to the path, 
    \item the cross-error $e_{c^{\bot}}$ perpendicular to the path, 
    \item the velocity error $e_v$, 
    \item and the directional error $e_d$. 
\end{compactenum}

The cross-error is the distance from the current position $x_t$ to the path $r^\mathrm{path}$. 
It can be split into the parallel part $e_{c^{\parallel}}$ and the perpendicular part $e_{c^{\bot}}$ according to the surface of the workpiece.
Here, perpendicular means the part that aligns with the surface normal of the workpiece, essentially displaying the height deviation of the robot from the path. This way, we can define separate weightings for both cross-error parts. 
Given the position of the closest control point $x^\mathrm{prior, path}_{m}$ according to the current position $x_t$, the parallel part of the cross-error can be computed as
\begin{equation}
    e_{c^{\parallel}}(s_t) = \left(x_{x,t} - x^\mathrm{prior, path}_{x, m}\right),
\end{equation}
while the perpendicular part is computed as 
\begin{equation}
    e_{c^{\bot}}(s_t) = \sqrt{ \left(x_{y,t} - x^\mathrm{prior, path}_{y, m}\right)^2 + \left(x_{z,t} - x^\mathrm{prior, path}_{z, m}\right)^2}.
\end{equation}
Note that this logic only applies to our task setup, where the x-direction always describes the lateral deviation from the path.
Both errors are transformed into rewards $r_{c^{\parallel}}$ and $r_{c^{\bot}}$ using transformation function $l(e)$.

In addition to that, we want to reward good velocity behavior according to the predefined velocity target $v_T^\mathrm{target}$. Using the current velocity $\dot{p}$ from the state $s$, the velocity is computed as 
\begin{align}
    e_v(s_t) = \abs{\norm{\dot{x}_t} - v_T^{\mathrm{target}}}.
\end{align}
We transform this error $e_v$ into a reward $r_v$ by using the linear function $l(e)$.

Since the information about the movement direction is lost when using the norm of the velocity $\norm{\dot{x}_t}$ for the reward $r_v$, we define an additional movement direction reward.
For this, we make use of the current movement direction $d_t = \frac{\dot{x}_t}{\norm{\dot{x}_t}} \in \mathbb{R}^3$ and the path direction $d^\mathrm{path} \in \mathbb{R}^3$ of the closest control point. The path direction $d^\mathrm{path}$ can be computed for every control point as the normed distance vector to the subsequent control point. Given these two directions, we can compute the directional error as their alignment angle
\begin{align}
    e_d(s_t) = \arccos \left( d_{t} \cdot d^\mathrm{path}\right).
\end{align}
The final direction reward $r_d$ can be computed using the piecewise linear function $l(e)$. Splitting the velocity and the direction into two rewards again enables independent reward weighting.

\fakepar{Force Tracking}
We define a force error based on the specified target force $F_N^\mathrm{target}$. For this, the sensed force is normed $\norm{F_t} = F_{N,t}$ and subtracted from the target
\begin{align}
    e_f(s_t) = \abs{F_{N,t} - F_N^\mathrm{target}}.
\end{align}
The force error $e_f$ is transformed to the reward $r_f$ using the linear-quadratic function $q(e)$. This transformation function emphasizes the reward for force values $F_{N,t}$ close to the target force $F_N^{target}$.

\fakepar{Reward Weighting}
Table \ref{tab:reward_weightings} shows the weights of the reward terms, while Tab. \ref{tab:reward_bounds} defines the boundary values for the piecewise error transformation functions. We distinguish between simulated and real hardware environments. We define wider reward boundaries for the hardware environment because this environment is generally more challenging. Both SAC and CHEQ are trained using the same set of parameters.

\begin{table*}[tb]
    \centering
    \begin{tabular}{|l|ccccccc|}
        \hline
        weights & $c_{trunc}$ & $c_{term}$ & $c_{c^{\bot}}$ & $c_{c^{\parallel}}$ & $c_{v}$ & $c_{d}$ & $c_{f}$ \\ \hline
        Value (sim.) & -1 & 0.1 & 0.1 & 0.05 & 0.3 & 0.15 & 0.4 \\
        Value (real) & -1 & 0.03 & 0.1 & 0.05 & 0.35 & 0.2 & 0.3 \\ \hline
    \end{tabular}
    \caption{Reward weightings for the simulated and real hardware environment.}
    \label{tab:reward_weightings}
\end{table*}

\begin{table*}[tb]
    \centering
    \begin{tabular}{|l|c|c|c|c|cc|}
        \hline
        reward & $r_{c^{\bot}}$ & $r_{c^{\parallel}}$ & $r_v$ & $r_d$ & $r_f$ & $r_f$ \\ \hline
        boundaries & $e_{max} \, [\si{\metre}]$ & $e_{max} \, [\si{\metre}]$ & $e_{max} \, [\si{\metre \per \second}]$ & $e_{max} \, [\si{\radian}]$ & $e_{max} \, [\si{\newton}]$ & $e_{mid} \, [\si{\newton}]$ \\ \hline
        Value (sim.) & 0.01 & 0.01 & 0.01 & 0.6 & 2.0 & 0.5 \\
        Value (real) & 0.015 & 0.03 & 0.03 & 0.6 & 4.0 & 1.5  \\ \hline
    \end{tabular}
    \caption{Reward transformation bounds for the simulated and real hardware environment.}
    \label{tab:reward_bounds}
\end{table*}

\subsection{Safety Limits}
\label{sec:app_safety_limits}
In the following, we describe the safety limitations of the polishing environment. A violation of one of the limitations will lead to truncation of the episode. We use four types of limits: positional, orientational, velocity and force limits.

\fakepar{Positional limitations}
\begin{figure}[tb]
    \centering
    \includegraphics[width=0.5\columnwidth, trim=0cm 0cm 0cm 0cm]{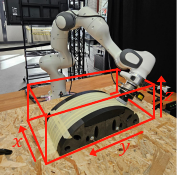}
    \caption{Positional safety limitations of the environment and the corresponding task space coordinate system.}
    \label{fig:pos_limits}
\end{figure}
In some applications, robotic arms are placed close to other facilities or stations of an automation process. Our positional limits restrict the arm movement inside a bounding box. We define the possible workspace of the robotic arm to be a cuboid, limiting the position $x=(x_x,x_y,x_z)$ of the end-effector close to the workpiece of the polishing process.
The workspace cuboid is defined as 
\begin{equation}
 \begin{aligned}
    &x_x \in [0.05, 0.15] \, [\si{\metre}], \\
    &x_y \in [-0.23, 0.57] \, [\si{\metre}] \text{, and} \\
    &x_z \in [0.0, 0.2] \, [\si{\metre}].
\end{aligned}   
\end{equation}
Here, the restriction of the $x_x$-position is tight compared to the other axes since the target path $r^\mathrm{path}$ is defined in the $yz$-plane. The asymmetric boundaries of the $x_y$-position arise because the origin of the y-axis is defined by the initial state of the polishing path. The z-axis has its origin on the table, and the origin of the x-axis is at the border of the workpiece.
Note that the positional limitations also passively restrict the robotic arm's joint displacements, so we do not define additional joint limits in our environment.
Figure \ref{fig:pos_limits} shows the positional limits.

\fakepar{Orientational limitations}
In addition to positional limits, we also restrain the end-effector orientation. Combined with the positional limits, this aims to avoid collision with the surroundings, particularly with the workpiece. We speak of collision when the robotic arm has non-planned contact with the environment, \ie if a part other than the end-effector has contact with the environment. 
In Euler angle representation [\si{\degree}], the orientational limits are defined as
\begin{equation}
 \begin{aligned}
    &\phi_x \in [-180, -110] \cup [110, 180], \\
    &\phi_y \in [-10, 10] \text{, and} \\
    &\phi_z \in [-10, 10].
\end{aligned}   
\end{equation}
Here, $\phi_x$ has a wider span compared to the other two rotation axes $\phi_y$ and $\phi_z$ since this rotation axis is used to perform the rotation movement of the end-effector.

\fakepar{Velocity limitations}
To ensure safe movement of the robotic arm, we restrict the end-effector velocity $\dot{x}$.
The limit of the velocity is defined as $\norm{\dot{x}} < \qty{0.5}{\metre \per \second}$. Anything above this threshold is considered dangerous for any setting of impedance gains.

\fakepar{Force limitations}
We aim to avoid high-force interactions between the end-effector and the workpiece. To yield this, we limit the maximum contact force of the force sensor to a certain threshold $F_N < \qty{25}{\newton}$. 
In addition to that, no contact with the environment is allowed when the end-effector position is below $z < \qty{0.01}{\metre}$. Detecting a contact force in this position means the end-effector has undesired contact with the workpiece table.

\subsection{Hyperparameters}
\label{sec:app_hyperparameters}
This section details the hyperparameters used for our experiments.

\fakepar{SAC Hyperparameters}
All SAC-specific hyperparameters for the simulated and hardware CHEQ agents, as well as the SAC variants, are described in Table \ref{tab:hyperparameter_sac}. To mitigate fluctuating $\lambda^\mathrm{RL}$ values on hardware, we choose a larger critic ensemble of \num{10} neural networks. For all algorithms, we include a random sampling phase for the first \num{15e3} steps where we sample the RL action uniformly random and do not update our agent. In this setting, we vary $\lambda^\mathrm{RL}$ between [0.2, 0.3] for the CHEQ agent.

\begin{table*}[tb]
\centering
\begin{tabular}{|l|ccc|}
    \hline
    Environment & \multicolumn{2}{|c|}{Simulation} & Real Hardware \\ \hline
    Parameter & Value (SAC) & \multicolumn{2}{|c|}{Value (SAC-Ensemble)} \\ \hline
    optimizer & ADAM & ADAM & ADAM \\
    $\eta$ & $1 \cdot 10^{-4}$ & $3 \cdot 10^{-4}$ & $3 \cdot 10^{-4}$ \\
    $\gamma$ & $0.99$ & $0.99$ & $0.99$ \\
    $\xi$ & $0.005$ & $0.005$ & $0.005$ \\
    $\abs{\mathcal{D}}$ & $10^6$ & $10^6$ & $10^6$ \\
    hidden-layers (critic NNs) & $[256, 256]$ & $[256, 256]$ & $[256, 256]$ \\
    hidden-layers (policy NN) & $[256, 256]$ & $[256, 256]$ & $[256, 256]$ \\
    \# of critic networks & $2$ & $5$ & $10$ \\
    batch-size & $256$ & $256$ & $256$ \\
    UTD & - & $1$ & $2$ \\
    non-linearity & ReLU & ReLU & ReLU \\
    $\alpha$ & $0.2$ & $0.2$ & $0.2$ \\
    $\alpha$-tuning & \text{True} & \text{True} & \text{True} \\
    initial random steps & $15000$ & $15000$ & $15000$ \\ \hline
\end{tabular}
\caption{Hyperparameters for the SAC agents.}
\label{tab:hyperparameter_sac}
\end{table*}

\fakepar{CHEQ Hyperparameters}
Table \ref{tab:cheq_hyperparams} provides the hyperparameters specific to the CHEQ algorithm.

\begin{table*}[tb]
    \centering
    \begin{tabular}{|l|cc|}
        \hline
        Parameter & Value (sim.) & Value (real) \\ \hline
        $\lambda_\mathrm{min}$ & 0.2 & 0.2 \\
        $\lambda_\mathrm{max}$ & 1.0 & 1.0 \\
        $u_\mathrm{min}$ & 0.02 & 0.015 \\
        $u_\mathrm{max}$ & 0.2 & 0.1 \\   
        horizon $T$ & 380 & 150 \\ \hline 
    \end{tabular}
    \caption{Hyperparameters of CHEQ used for the weight adaption function $\lambda^\mathrm{RL}$ for the hybrid agent.}
    \label{tab:cheq_hyperparams}
\end{table*}

As the control frequency $f^\mathrm{RC, real} = \qty{20}{\hertz}$ of the real hardware environment is lower than the control frequency of the simulation $f^\mathrm{RC, sim} = \qty{50}{\hertz}$, it takes less episodic steps to perform the task successfully. We thus choose different horizon lengths for the two environments. This results in less accumulated return, which alters the values of the Q-networks. Hence, the Q-networks show lower values in the hardware environment than in the simulation. The final return in simulation is approximately half the simulated return. Lower values result in lower standard deviations or total uncertainty values. Thus, lower uncertainty limits must be chosen for the hardware setup to achieve the same behavior. Since we focus on safe exploration in our hardware experiments, we have increased the upper limit $u_\mathrm{max}$ to achieve safer exploration behavior through lower $\lambda^\mathrm{RL}$ values.

\fakepar{Nominal Control Hyperparamters}
\label{sec:app/nominal/hyperparameters}

Table \ref{tab:hyperparameter_nominal} provides the hyperparameters for the nominal controller for the simulated and real environment.

\begin{table*}[tb]
\centering
\begin{tabular}{|l|ccc|}
    \hline
    Environment & \multicolumn{2}{|c|}{Simulation} & Real Hardware \\ \hline
    Parameter & baseline & \multicolumn{2}{|c|}{suboptimal} \\ \hline
    $a_r$ [\si{\metre}] & 0.016 & 0.2 & 0.015 \\
    $\Delta x^\mathrm{indent}$ [\si{\metre}] & 0.012 & 0.015 & 0.002 \\
    spacing [\si{\metre}] & $5 \cdot 10^{-4}$ & $5 \cdot 10^{-4}$ & $5 \cdot 10^{-4}$ \\ \hline
\end{tabular}
\caption{Hyperparameter choices for the nominal controller for the simulation and the real hardware environment. In addition, we define the suboptimal controller that is used for CHEQ. As optimal gains for the real process are unknown, we label the parameters for the real process as suboptimal as well.}
\label{tab:hyperparameter_nominal}
\end{table*}

The spacing parameter describes the equidistant spacing between neighbouring control points used during path $r^\mathrm{path}$ creation. This setup consistently provides $\sim 50$ control points as close control points $P^\mathrm{close}$. Further, note that we use a low value for the indentation $\Delta x^\mathrm{indent}$ in the real environment. This is due to the fact that the movement of the real robot is more cumbersome than in the simulated environment. Setting a lower indentation leads to a stronger movement in the direction of the path. 

\fakepar{Baseline and Impedance Gains}
To define our baseline control (\textit{Nominal (tuned)}), we perform hyperparameter tuning with Bayesian Optimization to find the optimal static impedance gains for the baseline controller in simulation.
As we expect the important impedance gains to be $k_y$ and $k_z$, we use these values and the damping factor $\zeta$ for the search space. We keep the other impedance gains fixed at values that are high enough to perform the given movement by the nominal controller. We choose $k_x = \qty{500}{\newton \per \metre}, k_{\phi_x} = k_{\phi_y} = k_{\phi_z} = \qty{500}{\newton \per \radian}$. For a better comparison of the learning capabilities, we define the same bounds for the search space as provided by the action space of the RL agent. Thus the search space is spanned as $k_y \in [50, 200] \, [\si{\newton \per \metre}]$, $k_z \in [30, 130] \, [\si{\newton \per \metre}]$, and $\zeta \in [0.8, 1.2]$. We perform a total of \num{2000} runs and rank them according to the predefined score of the task
\begin{equation}
    \mathrm{score} = w_c \cdot \frac{N_\mathrm{con}}{N_\mathrm{tot}} + w_w \cdot \frac{n_\mathrm{wipes}}{7} - w_f \cdot \overline{\Delta F_N} - w_v \cdot \overline{\Delta v_T},
    \label{eq:bo_score}   
\end{equation}
where $N_\mathrm{tot}, \, N_\mathrm{con}$ are the numbers of total episodic steps and steps with contact to the workpiece, $n_\mathrm{wipes}$ describes the number of wiped wiping points, and $\overline{\Delta F_N}, \, \overline{\Delta v_T}$ are the average deviations of force and velocity to the specified targets. The constant values at the beginning of each score term are the weighting factors. Since we favour runs that polish over the complete task space in the predefined horizon $T$, we set a relatively high weighting to the wiping and contact term. The weighting factors are configured as $w_c = 0.18$, $w_w = 0.52$, $w_f = 0.03$, and $w_v = 0.27$. In the end, we choose the run resulting in the highest score as the impedance gains of the \textit{Nominal (tuned)} controller, that could be found after $\sim$ \num{500} runs. For the control prior used for the CHEQ agent, we further define a suboptimal controller (\textit{Nominal (untuned)}) determined by rounding up the values.

\begin{table*}[ht]
    \centering
    \begin{tabular}{|l|ccc|ccc|c|}
        \hline
        Controller & $k_x$ & $k_y$ & $k_z$ & $k_{\phi_x}$ & $k_{\phi_y}$ & $k_{\phi_z}$ & $\zeta$ \\ \hline
        \textit{Nominal (untuned)} & $500$ & $160$ & $50$ & $500$ & $500$ & $500$ & $1.0$ \\
        \textit{Nominal (tuned)} & $500$ & $107$ & $68$ & $500$ & $500$ & $500$ & $0.9562$ \\
        \textit{Nominal (hardware, untuned)} & $800$ & $1085$ & $900$ & $150$ & $80$ & $25$ & $1.0$ \\ 
        \hline
    \end{tabular}
    \caption{Definition of impedance gains for the fixed gain nominal controllers.}
    \label{tab:gains_baseline}
\end{table*}

For the partially tuned nominal controller \textit{Nominal (tuned section-wise)}, we divide the workpiece into \num{5} distinct sections. 
The division of the task is based on the current $y$-position of the end-effector.
We find the gains for the partitioned controller by iteratively tuning with Bayesian Optimization for each section and using the best-found gains for the previous sections. Table \ref{tab:partition_impedance} provides the final impedance gains configurations using this iterative method.

\begin{table*}[ht]
    \centering
    \begin{tabular}{|c|c|c|c|}
        \hline
        partition & $k_y \, [\si{\newton \per \metre}]$ & $k_z \, [\si{\newton \per \metre}]$ & $\zeta$ \\ \hline
        \textbf{S1} & $98$ & $59$ & $\sim 0.9445$ \\ 
        \textbf{S2} & $91$ & $69$ & $\sim 0.8563$ \\
        \textbf{S3} & $108$ & $71$ & $\sim 0.9452$ \\
        \textbf{S4} & $137$ & $72$ & $\sim 1.0294$ \\
        \textbf{S5} & $101$ & $65$ & $\sim 1.1946$ \\ \hline
    \end{tabular}
    \caption{Impedance profile for the \textit{Nominal (tuned section-wise)} controller.}
    \label{tab:partition_impedance}
\end{table*}
%
%
%
%
%
\section{Additional Simulation Results}              
This section describes additional results.

\subsection{Comparison of Tuned Baselines with Trained CHEQ Agent}
\label{sec:app_comp_baseline_cheq}

\begin{figure*}[tb]
     \centering
        \begin{subfigure}[b]{\textwidth}
        \begin{flushright}
        \includegraphics[clip, trim=0.1cm 0.1cm 0.1cm 0.1cm]{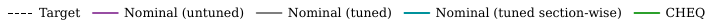}
        \end{flushright}
     \end{subfigure}
     \begin{subfigure}[b]{0.32\textwidth}
         \centering
        \includegraphics[clip, trim=0cm 0.25cm 0cm 0.25cm]{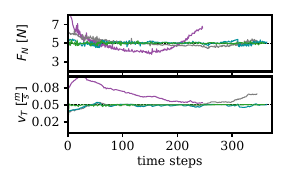}
         \caption{}
         \label{fig:fv_behavior_raw_tuned_partial_cheq}
     \end{subfigure}
     \begin{subfigure}[b]{0.32\textwidth}
         \centering
        \includegraphics[clip, trim=0cm 0.25cm 0cm 0.25cm]{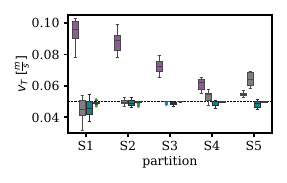}
         \caption{}
         \label{fig:v_box_raw_tuned_partial_cheq}
     \end{subfigure}
     \begin{subfigure}[b]{0.32\textwidth}
         \centering
        \includegraphics[clip, trim=0cm 0.25cm 0cm 0.25cm]{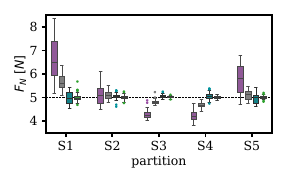}
         \caption{}
         \label{fig:f_box_raw_tuned_partial_cheq}
     \end{subfigure}
     \begin{subfigure}[b]{0.32\textwidth}
         \centering
        \includegraphics[clip, trim=0cm 0.25cm 0cm 0.25cm]{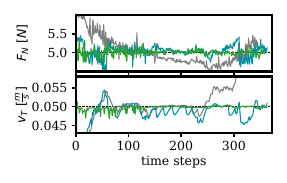}
        \caption{}
        \label{fig:fv_behavior_tuned_partial_cheq}
     \end{subfigure}
     \begin{subfigure}[b]{0.32\textwidth}
         \centering
        \includegraphics[clip, trim=0cm 0.25cm 0cm 0.25cm]{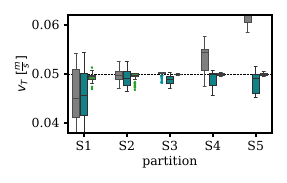}
        \caption{}
         \label{fig:v_box_tuned_partial_cheq}
     \end{subfigure}
     \begin{subfigure}[b]{0.32\textwidth}
         \centering
        \includegraphics[clip, trim=0cm 0.25cm 0cm 0.25cm]{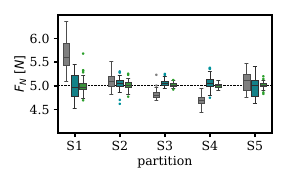}
        \caption{}
         \label{fig:f_box_tuned_partial_cheq}
     \end{subfigure}
     \caption{Comparison of the trained CHEQ agent with the nominal control baselines.}
     \label{fig:app_raw_fix_partial_cheq}
\end{figure*}
For a better comparison, we show the results of our trained CHEQ agent in comparison with the three nominal control baselines \textit{Nominal (untuned)}, \textit{Nominal (tuned)} and \textit{Nominal (tuned section-wise)} in Fig. \ref{fig:app_raw_fix_partial_cheq}. In the first row, we compare the untuned controller, the tuned fixed gain controller, the section-wised tuned controller and our trained CHEQ agent. The second row shows the same comparison but leaves out the untuned controller to enable a better comparison. We compare the force and velocity behavior over one full episode in the subfigures \ref{fig:fv_behavior_raw_tuned_partial_cheq} and \ref{fig:fv_behavior_tuned_partial_cheq}. The force and velocity boxplots for the five workpiece sections are shown in the subfigures \ref{fig:v_box_raw_tuned_partial_cheq}, \ref{fig:f_box_raw_tuned_partial_cheq}, \ref{fig:v_box_tuned_partial_cheq}, and \ref{fig:f_box_tuned_partial_cheq}.

\section{Additional Real-World Results}   
\label{sec:app_real_add_results}
This section describes additional experimental details and results for the hardware experiments.

\fakepar{Real-world RL Challenges and Solutions}
\label{sec:app_extensions_real_rl}
Additional challenges need to be considered to achieve stable training on hardware. We found (i) control frequency, (ii) sensor noise and (iii) fluctuations in $\lambda$ to be most important.

Simulation frameworks for RL, such as MuJoco~\citep{todorov_mujoco_2012}, specify a control frequency, which determines the actual control frequency $f^\mathrm{RC}$, \ie the time between control actions as well as the waiting time $\Delta t_{(a_t, s_{t+1})}$ between applying an action $a_t$ and collecting the next state $s_{t+1}$. These simulations assume no time spent on communication, action computation, or gradient updates, \ie $\Delta t_{(a_t, s_{t+1})}= (f^\mathrm{RC})^{-1}$. These assumptions are unrealistic in the real world, and the waiting time can only use up some fraction of the control frequency. This poses an inherent dilemma, as a faster control frequency can only be achieved with a lower waiting time. Sufficient waiting times, however, are critical for allowing the RL agent to observe the consequences of its actions.
To balance hardware limitations with the need for adequate waiting time, we selected a control frequency of \qty{20}{\hertz} and a waiting time of \qty{0.033}{\second}. In simulation, a higher control frequency of \qty{50}{\hertz} allowed for improved responsiveness to force changes.
In addition, we defined a split actor and learner setup, similar to \citet{luo_serl_2024}, to ensure that gradient updates do not lower the control frequency. This also allows the environment to reset parallel to the gradient steps.

The second challenge is the measurement of noise in the robot state and the force-torque sensor. In the latter, we encountered drift and motion-dependent sensor noise. This high noise level, combined with the low control frequency, complicates the training on hardware. We applied a low-pass filter with a cut-off frequency of \qty{35}{\hertz}. This simplified the optimization landscape and enabled the agent to focus on aspects within control.

In our simulation study, we found one critical aspect specific to the CHEQ algorithm and the VIC task. In the initial training phase, the weighting factor fluctuates greatly between time steps, resulting in highly fluctuating IC gains. In simulation, this does not pose a problem. On real hardware, however, the gain fluctuations lead to a chattering end-effector motion. This doubly affects the training as it complicates the control and increases the motion-dependent force sensor noise. To mitigate this, we use an average of \num{10} time steps to compute our uncertainty measurement, leading to a smoother weighting. 

Combining the abovementioned challenges, we find that data collected on our hardware setup is less reliable than in simulation. This can lead to unstable training. However, CHEQ provides a mechanism to mitigate this by slowing the curriculum. Thus, for our hardware experiments, we set the uncertainty limits to be more conservative. This reduces the agency of the RL agent in the beginning and leads to more stable, even safer, albeit slower, learning.

\begin{figure*}[tb]
    \centering
    \begin{subfigure}[b]{\textwidth}
        \begin{flushright}
        \includegraphics[clip, trim=0.1cm 0.1cm 0.1cm 0.1cm]{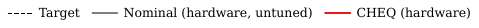}
        \end{flushright}
    \end{subfigure}
    \begin{subfigure}[b]{0.32\textwidth}
        \centering
        \includegraphics[clip, trim=0.25cm 0cm 0.25cm 0cm]{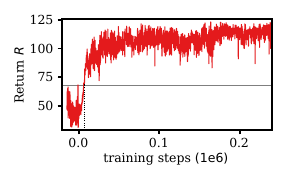}
        \caption{Initial training return.}
        \label{fig:cheq_rr_rollout_return_zoom}
    \end{subfigure}
    \begin{subfigure}[b]{0.32\textwidth}
        \centering
        \includegraphics[clip, trim=0.25cm 0cm 0.25cm 0cm]{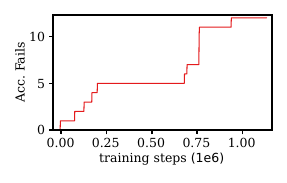}
        \caption{Accumulated fails (log scale).}
        \label{fig:fails_cheq_rr}
    \end{subfigure}
    \begin{subfigure}[b]{0.32\textwidth}
        \centering
        \includegraphics[clip, trim=0.25cm 0cm 0.25cm 0cm]{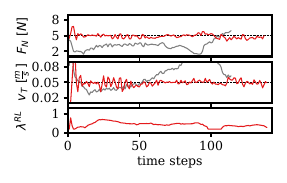}
        \caption{Late evaluation: $\sim 930000$}
        \label{fig:fvw_cheq_rr_late}
    \end{subfigure}
    \caption{We show a close look at the initial training return in Fig. (\subref{fig:cheq_rr_rollout_return_zoom}). Further, the accumulated training fails are shown in Fig. (\subref{fig:fails_cheq_rr}). Figure (\subref{fig:fvw_cheq_rr_late}) shows the force and velocity behavior of an evaluation run after longer training ($\sim 930000$ training steps).}
    \label{fig:cheq_rr_appendix}
\end{figure*}

\fakepar{Additional Hardware Results}
We show detailed hardware results in Fig. \ref{fig:cheq_rr_appendix}. The return distribution of the early training phases shows that the return rises fast and surpasses the control prior (see Fig. \ref{fig:cheq_rr_rollout_return_zoom}). The complete progression of training failures can be found in Fig. \ref{fig:fails_cheq_rr}. We further show the force, velocity and weight behavior of an evaluation run after more training steps (see Fig. \ref{fig:fvw_cheq_rr_late}). This underlines our findings that the agent learns to act with a higher weighting $\lambda^\mathrm{RL}$ but does not further improve the polishing behavior.

\end{document}